\documentclass[sigconf, natbib]{acmart}
\usepackage{enumitem}
\usepackage{multirow}
\usepackage{tabularx} 
\usepackage{booktabs}  
\usepackage{arydshln} 
\usepackage{siunitx} 
\usepackage{stfloats}
\usepackage{array}
\usepackage{graphicx}
\usepackage{adjustbox}
\usepackage{xcolor}
\usepackage{tcolorbox}
\usepackage{pifont}
\usepackage{listings} 

\usepackage[ruled,linesnumbered]{algorithm2e}  

\usepackage{amsmath}
\usepackage{amssymb}

\usepackage{bm}

\usepackage{float}

\usepackage{subcaption}

\definecolor{LightBlue}{RGB}{173,216,230}

\AtBeginDocument{%
  }

\copyrightyear{2025}
\acmYear{2025}
\setcopyright{cc}
\setcctype{by-nc}
\acmDOI{10.1145/3746252.3761210}

\acmConference[CIKM '25] {Proceedings of the 34th ACM International Conference on Information and Knowledge Management}{November 10--14, 2025}{Seoul, Republic of Korea}
\acmBooktitle{Proceedings of the 34th ACM International Conference on Information and Knowledge Management (CIKM '25), November 10--14, 2025, Seoul, Republic of Korea}
\acmISBN{979-8-4007-2040-6/2025/11}

\begin{document}

\title{Improved Personalized Headline Generation via Denoising Fake Interests from Implicit Feedback}

\author{Kejin~Liu}
\authornote{Both authors contributed equally to this work.}
\authornotemark[3]
\orcid{0009-0008-8710-6575}
\affiliation{%
  \institution{Henan Institute of Advanced Technology, Zhengzhou University}
  \state{Zhengzhou}
  \country{China}
} 
\email{liukejin@gs.zzu.edu.cn}

\author{Junhong~Lian}
\authornotemark[1]
\authornotemark[3]
\authornotemark[6]
\orcid{0000-0002-2922-1216}
\affiliation{%
  \institution{Institute of Computing Technology, Chinese Academy of Sciences}
  \state{Beijing}
  \country{China}
} 
\email{lianjunhong23s@ict.ac.cn}

\author{Xiang~Ao}
\authornote{Corresponding author.}
\authornote{State Key Laboratory of AI Safety, Institute of Computing Technology (ICT), Chinese Academy of Sciences (CAS).}
\authornote{Henan Institute of Advanced Technology, Zhengzhou University, Zhengzhou, China.}
\authornotemark[6]
\orcid{0000-0001-9633-8361}
\affiliation{%
  \institution{Institute of Computing Technology, Chinese Academy of Sciences}
  \state{Beijing}
  \country{China}
}
\email{aoxiang@ict.ac.cn}

\author{Ningtao~Wang}
\orcid{0009-0005-6577-5047}
\affiliation{%
  \institution{Independent Researcher}
  \state{Hangzhou}
  \country{China}
}
\email{ntwang25@gmail.com}

\author{Xing~Fu}
\orcid{0000-0002-3536-2779}
\affiliation{%
  \institution{Independent Researcher}
  \state{Hangzhou}
  \country{China}
}
\email{fux008@gmail.com}

\author{Yu~Cheng}
\orcid{0000-0001-5469-3509}
\affiliation{%
  \institution{Independent Researcher}
  \state{Hangzhou}
  \country{China}
}
\email{yu.cheng.info@gmail.com}

\author{Weiqiang~Wang}
\orcid{0000-0002-6159-619X}
\affiliation{%
  \institution{Independent Researcher}
  \state{Hangzhou}
  \country{China}
}
\email{wang.weiqiang@gmail.com}

\author{Xinyu~Liu}
\authornote{High Performance Computer Research Center, ICT, CAS.}
\authornote{The authors are also with the University of Chinese Academy of Sciences, CAS.}
\orcid{0009-0001-9481-5758}
\affiliation{%
  \institution{Institute of Computing Technology, Chinese Academy of Sciences}
  \state{Beijing}
  \country{China}
}
\email{liuxinyu@ict.ac.cn}

\renewcommand{\shortauthors}{Kejin Liu, Junhong Lian, Xiang Ao, Ningtao Wang, Xing Fu, Yu Cheng, Weiqiang Wang and Xinyu Liu}

\begin{abstract}
  Accurate personalized headline generation hinges on precisely capturing user interests from historical behaviors. However, existing methods neglect personalized-irrelevant click noise in entire historical clickstreams, which may lead to hallucinated headlines that deviate from genuine user preferences.
  In this paper, we reveal the detrimental impact of click noise on personalized generation quality through rigorous analysis in both user and news dimensions.
  Based on these insights, we propose a novel \textbf{\underline{P}}ersonalized \textbf{\underline{H}}eadline \textbf{\underline{G}}eneration framework via \textbf{\underline{D}}enoising Fake Interests from \textbf{\underline{I}}mplicit \textbf{\underline{F}}eedback~(\textbf{PHG-DIF}). 
  PHG-DIF first employs dual-stage filtering to effectively remove clickstream noise, identified by short dwell times and abnormal click bursts, and then leverages multi-level temporal fusion to dynamically model users’ evolving and multi-faceted interests for precise profiling.
  Moreover, we release \textbf{DT-PENS}, a new benchmark dataset comprising the click behavior of $1,000$ carefully curated users and nearly $10,000$ annotated personalized headlines with historical dwell time annotations.
  Extensive experiments demonstrate that PHG-DIF substantially mitigates the adverse effects of click noise and significantly improves headline quality, achieving state-of-the-art~(SOTA) results on DT-PENS. 
  Our framework implementation and dataset are available at \url{https://github.com/liukejin-up/PHG-DIF}.
\end{abstract}



\maketitle

\section{Introduction}
\label{sec:intro}
Personalized headline generation has emerged as a pivotal strategy for enhancing user engagement on news platforms~\cite{ao2021pens}. Prevailing methods typically condense user profiles from historical clickstreams into compact representations for personalization~\cite{zhang2022personalized, ao2023put, lian2025panoramic}. Despite the undeniable success of these personalized methods and continued progress, they predominantly rely on users' entire click history~\cite{yang2023fact, song2023general, tan2024enhancing}, overlooking a crucial characteristic of the clickstream: the inherent uncertainty of user click behaviors~\cite{jiang2024time}.

\begin{figure}[t]
    \centering
    \includegraphics[width=0.92\linewidth]{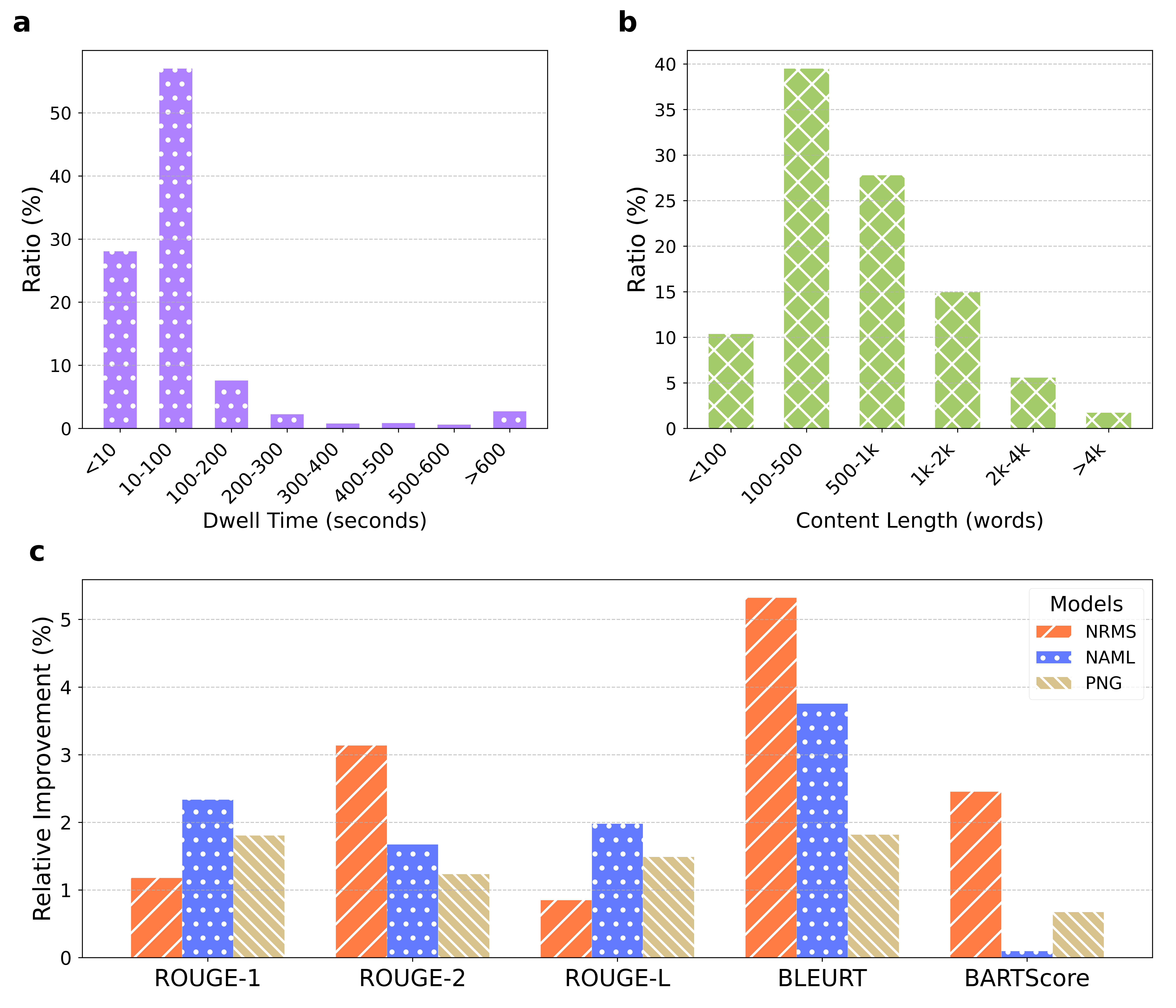}
    \caption{
    Fig.~(a) and (b) present the ratio distribution of dwell times on the click history and news content length for a certain MSN user, respectively.
    Fig.~(c) illustrates the relative improvement in evaluation after denoising.}
    \label{fig:intro}
    \vspace{-1em}
\end{figure}

User click behaviors are influenced by a multitude of uncertain factors, extending beyond direct interest reflections~\cite{xie2021deep, jiang2024time}. We regard clicks that are unrelated to personalization as ``click noise'' within historical clickstreams. 
Empirical analysis\footnote{Analysis based on PENS~\cite{ao2021pens}. We randomly sampled users (e.g., \textsf{U362229} in Figure~\ref{fig:intro}) to calculate distributions of click history and corresponding news content length.} shows that dwell time, the duration users spend reading news articles, effectively indicates click noise. 
A simple case study with a randomly selected MSN user reveals that $28.08\%$ of clicked news exhibit dwell times less than 10 seconds, while merely $10.39\%$ involve news content exceeding 100 words, as shown in Figure~\ref{fig:intro}a-\ref{fig:intro}b. Cognitive studies suggest even professional speed readers would struggle to fully comprehend 100-word content in under 10 seconds~\cite{rayner2016so}. 
Moreover, transient news events (e.g., 2020 U.S. elections) induce temporal click surges containing incidental clicks from contextually uninterested users attracted by platform recommendations. 
By applying a simple, rule-based filtering method to exclude clicks with dwell times under 10 seconds and the top $0.1\%$ of click-through rates during specific user impression log periods, we observed a significant improvement in personalization, as shown in Figure~\ref{fig:intro}c.

Therefore, we attribute click noise in personalized user profiles to \textbf{the user dimension} and \textbf{the news dimension}. Users may rapidly exit from news due to misclicks or misleading headlines, resulting in clicks unrelated to their interests and forming click noise in the user dimension. In contrast, click noise in the news dimension typically arises from transient news events, which trigger a surge of clicks from non-specific users.
These users are influenced by platform recommendations, not genuine interest. Both types of click noise hinder the accurate capture of user preferences, causing generated headlines to diverge from users' true interests.
However, tackling historical click noise originating from both dimensions simultaneously remains challenging. A key challenge is the complexity of user click behaviors and the dynamic evolution of user interests, which complicates precise user profiling. Additionally, the absence of user historical dwell time data in existing personalized headline generation benchmark restricts further evaluation.

To remedy these challenges, we propose \textbf{PHG-DIF}, a novel \textbf{\underline{P}}ersonalized \textbf{\underline{H}}eadline \textbf{\underline{G}}eneration framework via \textbf{\underline{D}}enoising Fake Interests from \textbf{\underline{I}}mplicit \textbf{\underline{F}}eedback.
PHG-DIF captures genuine user interests through a dual-filtering strategy, which filters out potential interference at both news-level and time-level. Concurrently, we capture users' multi-faceted preferences via dedicated modules for \emph{Instantaneous Preference Learning~(IPL)}, \emph{Interest Evolution Analysis~(IEA)}, and \emph{Stable Interest Mining~(SIM)}. These preferences are fused by multi-granular dynamic aggregation into a unified user representation that is subsequently injected at each decoding step of a breaking-news-aware pointer generator, thereby balancing factual accuracy with personalized headlines.
Furthermore, we introduce \textbf{DT-PENS}, an extended benchmark derived from PENS~\cite{ao2021pens}. DT-PENS includes complete dwell-time logs and nearly $10,000$ human-validated personalized headlines for $1,000$ carefully curated users, providing a comprehensive resource for mitigating historical click noise and personalized modeling.

The main contributions of this paper are summarized as follows:
\begin{itemize}
    \item We propose a novel framework, PHG-DIF, which captures genuine user interests through dual-filtering and enhances personalization by multi-faceted user modeling.
    \item We introduce DT-PENS, an extended benchmark with nearly $10,000$ personalized headlines for $1,000$ carefully curated users annotated with historical dwell time, enabling more robust evaluation of personalized headline generation.
    \item Extensive experiments demonstrate PHG-DIF substantially mitigates the impact of click noise and significantly improves the quality of generated personalized headlines, achieving SOTA performance on DT-PENS benchmark.
\end{itemize}

\section{Related Work}
\label{sec:related_work}
News platforms increasingly rely on automated headline generation to enhance user engagement and content distribution efficiency~\cite{ao2021pens}. Early studies predominantly followed a content-compression paradigm, where headlines were either extracted or generated from news content to summarize the main idea, essentially forming a specialized instance of text summarization~\cite{rush2015neural, nallapati2016abstractive,luo2019reading,li2022news}. 
Unified headlines often fail to accommodate individual user preferences, especially when readers focus on vastly different aspects of the same event, thus limiting user engagement~\cite{ao2021pens}. Consequently, automated news headline generation is undergoing a paradigm shift from generic summarization towards personalization, a necessary transition driven by news platforms' pursuit of refined user operations and higher user retention.

PENS~\cite{ao2021pens} was the first to formally define personalized news headline generation and introduced a large-scale public benchmark for offline evaluation. This spurred the development of various representative methods that encode user interests from click histories and integrate them into the headline generation process~\cite{zhang2022personalized,ao2023put,tan2024enhancing}. These approaches typically encode entire historical clickstreams into user-interest embedding generators, significantly outperforming generic models on metrics such as ROUGE and BLEU. 
GTP~\cite{song2023general} further decomposed the generation process into a generic headline generation stage followed by personalized refinement. Concurrently, FPG~\cite{yang2023fact} employed contrastive learning to constrain factual consistency, thus avoiding misleading attention-grabbing headlines.

The core of personalized generation lies in accurately modeling user interests. The field of recommendation systems has witnessed the evolution of several paradigms, from static vectors~\cite{kim2003learning} to sequential attention mechanisms~\cite{zhou2018din,zhou2019dien}, and subsequently to hierarchical interest modeling~\cite{qi2021hierec,qian2023hutcrs}. 
Research on personalized headline generation has also begun exploring dimensions beyond content, including style, tone, and entity preferences. LaMP~\cite{salemi2024lamp} introduced a personalization task leveraging authorial styles to achieve reader-side style matching, exploring personalization from a novel perspective. SCAPE~\cite{lian2025panoramic} explicitly modeled user preferences across both content and stylistic dimensions, integrating short- and long-term interests to enhance fine-grained dynamic profiling.

However, limitations persist as most existing methods~\cite{ao2021pens, ao2023put} still rely on the ``click equals interest'' assumption for user modeling. Such an assumption often leads to the inadvertent incorporation of noise from users' historical clickstreams into their profiles. As a result, these profiles can deviate from genuine preferences. 
Indeed, studies show that noisy implicit feedback significantly undermines model robustness in news recommendation~\cite{wang2021denoising, jiang2024time, xie2023reweighting}. Furthermore, news content is highly time-sensitive, and user interests evolve rapidly. These factors make it challenging to accurately capture dynamic, multi-dimensional preferences. Filtering noise from user clicks and modeling their dynamic interests are key steps toward precise personalized headline generation. Nevertheless, this potential remains largely unexplored.

\section{Problem Formulation}
For any user $u$, the click history $C_u$ is defined as an ordered sequence of $N$ interactions: 
$C_u = [(h_1^{u}, t_1^{u}), (h_2^{u}, t_2^{u}), \dots, (h_N^{u}, t_N^{u})]$, 
where $h_i^{u}$ denotes the headline of a news article clicked by user $u$, and $t_i^{u}$ represents the corresponding dwell time.
Our goal is to construct a valid click noise-irrelevant user click history $C_u^v = [(h_{k_1}^{u}, t_{k_1}^{u}), (h_{k_2}^{u}, t_{k_2}^{u}), \dots, (h_{k_n}^{u}, t_{k_n}^{u})] \subseteq C_u$ that captures $u$'s genuine interests. 
Then, given a candidate news article with original headline $h_x$ and body content $b_x$, we subsequently generate a personalized headline $Y_u^v$ for user $u$ based on $C_u^v$.

\section{Our Framework}
\begin{figure*}[t]
    \centering
    \includegraphics[width=0.8\linewidth]{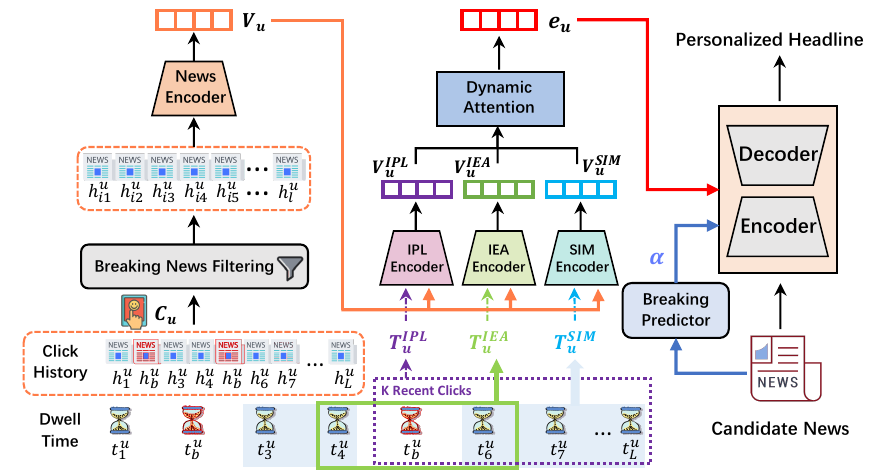}
    \caption{Overview of the proposed PHG-DIF framework.}
    \label{fig2}
    \vspace{-1em}
\end{figure*}

An overview of the PHG-DIF framework is shown in Figure~\ref{fig2}.

\subsection{User Modeling with Dual-Filtering}\label{subsec:time-encoder}
\paragraph{News-Level Filtering}
We first construct breaking news candidates by selecting the top-$M$ most clicked news headlines from the PENS dataset, forming set $B = \{h^\text{b}_j\}_{j=1}^M$, which serves as positive samples for training our breaking news predictor. 
For each user's click history $C_u = \{h^u_i\}_{i=1}^L$ where $L$ denotes the total clicked news count, we mask dwell times ($t^u_i \leftarrow 0$) for $h^u_i \in B$, obtaining filtered sequence $H_u = (h_{i1}^u, \ldots, h_l^u)$.
\paragraph{Time-Level Filtering}
Building upon the news-level filtered click history $H_u$, we further refine modeling of user's dwell time-based reading pattern. We align position indices between the filtered news sequence $H_u$ and original dwell times sequence $T_u = (t_1^{u}, ..., t_L^{u})$ via zero-padding. 
Each headline in $H_u$ is first represented by its word embeddings and then encoded using a news encoder, which employs attention mechanism to generate the news representation~$V_u$. 

To effectively model the interaction between the user's historical dwell times and $V_u$, we introduce three specialized time-aware encoders. These encoders operate at multiple granularities to capture different aspects of user preference: 

\textit{(1)~Instant Preference Learning~(IPL):} 
We first capture the user's immediate interests by focusing on the most recent $K$ click histories and their associated dwell times by IPL.
Specifically, the dwell time $T_u^{\text{IPL}}$ for the $i$-th item is defined as follows:
\begin{align}
T_u^{\text{IPL}} =
\begin{cases}
t_i, & i \in [L-K+1, L], \\
0, & \text{otherwise},
\end{cases}
\label{Equation 3}
\end{align}
where $t_i$ is the dwell time for the $i$-th item, $L$ is total click histories. The user's instant preference representation $v_u^{\text{IPL}}$ is then computed as follows:
\begin{align}
v_u^{\text{IPL}} =  w^{\text{IPL}} \cdot V_u,
\label{Equation 4}
\end{align}
where $w^{\text{IPL}}$ is learnable weight vector and $V_u$ is the embedding matrix of the user's click histories.

By focusing on the most recent interactions, IPL effectively captures and prioritizes the user's current interests. For instance, even if a user historically favored the Golden State Warriors, their present focus on the Miami Heat is reflected in IPL, aligning with their current preferences.

\textit{(2)~Interest Evolution Analysis~(IEA):} 
Then, we introduce IEA to model the dynamic changes in user interests based on their click history and dwell times within a certain time window $n$. 
By emphasizing temporal interaction patterns, IEA detects emerging interests, adapting to both abrupt and gradual changes in user behavior.
The dwell time $T_u^{\text{IEA}}$ for the $i$-th item and the user's evolving interest representation $v_u^{\text{IEA}}$ are defined as:  
\begin{align}
T_u^{\text{IEA}} =
\begin{cases}
t_i, & t_i \in [T - n, T], \\
0, & \text{otherwise},
\end{cases}
\label{Equation 5}
\end{align}
\begin{align}
v_u^{\text{IEA}} = w^{\text{IEA}} \cdot V_u,
\label{Equation 6}
\end{align}
where $T$ is the current time, $n$ is the length of the time window, and $w^{\text{IEA}}$ is learnable weight vector.

\textit{(3)~Stable Interest Mining~(SIM):} 
In SIM, we identify a user’s long-term interests by analyzing news articles with consistently high dwell times across their click history. The threshold for stable interest is defined as the mean dwell time. For instance, if a user frequently engages with health and economics topics, these areas are considered their stable interests. The stable interest signals are formally defined as follows:
\begin{align}
T_u^{\text{SIM}} =
\begin{cases}
t_i, & t_i > \textit{mean}(t), \\
0, & \text{otherwise},
\end{cases}
\label{Equation 7}
\end{align}
\begin{align}
v_u^{\text{SIM}} = w^{\text{SIM}} \cdot V_u,
\label{Equation 8}
\end{align}
where $\textit{mean}(t)$ is the average dwell time across all interactions, and $w^{\text{SIM}}$ is learnable weight vector.

\subsection{Multi-Granular Dynamic Aggregation}  
We aggregate the embeddings from \textit{IPL}, \textit{IEA}, and \textit{SIM} using a dynamic attention mechanism, enabling multi-level cross-time fusion to capture the user's multidimensional interest vector $e_u$. The matrices $w^{\text{i}}$ ($i \in \{\mathrm{IPL}, \mathrm{IEA}, \mathrm{SIM}\}$) from the previous section are Min-Max scaled from $T_u^{\text{i}}$.
\begin{align}
e_u = \text{DynAttn}[v_u^{\text{IPL}}, v_u^{{\text{IEA}}}, v_u^{{\text{SIM}}}].
\label{Equation_Dyn}  
\end{align}

Here, $\textit{DynAttn}$ dynamically adjusts the weights of multi-granular interests, producing a unified representation $e_u$ that reflects the user's true preferences. 

\subsection{Breaking-News-Aware Generator}
The candidate news may be potential breaking news, which tends to be inherently more eye-catching. Users typically prefer objective headlines in such cases, requiring less personalization. 
A feasible approach is to train a BERT‐based classifier $B_{\psi}$ on the top–$M\%$ click-through headlines~\citep{devlin-etal-2019-bert}, as breaking news often revolves around similar topics, such as the lives and careers of public figures.
During the headline generation phase, candidate news body $b_x$ is processed by an encoder to obtain its dense representation $v$. Subsequently, a breaking news predictor utilizes $v$ to yield a probability:
\begin{align}
\alpha = B_{\psi}(v)\in[0,1],
\end{align}
where a higher $\alpha$ indicates a greater likelihood that the article is breaking news.
For breaking news, the headline should emphasize facts, while other news can be personalized to user preferences.

Inspired by~\citet{ao2021pens, ao2023put}, we instantiate the personalized generator $G_{\theta}$ as a pointer-network decoder that conditions on both the encoded news representation $v$ and the user interest vector $e_u$.  
At each decoding step $t$, the decoder state $s_t$ attends over the encoder hidden states $h_j$, producing an attention distribution $a_{t,j}$ and a context vector $c_t=\sum_j a_{t,j}h_j$.  
Following \citet{See_Liu_Manning_2017}, the final vocabulary distribution is a convex combination of the generator distribution $P_v(\cdot)$ and the copy distribution induced by $a_t$:
\begin{align}
P(w) &= \lambda_t \, P_{v}(w) + (1 - \lambda_t) \sum_{j:w_j=w} a_{t,j}, \label{eq:ptr}\\
\lambda_t &= \sigma\!\bigl(W_{\lambda}[c_t; s_t; \alpha\,e_u] + b_{\lambda}\bigr),\label{eq:gate}
\end{align}
where $W_{\lambda}$ and $b_{\lambda}$ are learnable parameters and $\sigma(\cdot)$ is the sigmoid function.  
The gating term $\lambda_t$ is modulated by $\alpha$:  
when $\alpha\!\to\!1$ (breaking news), $\lambda_t$ tends to favor factual copying from $b_x$; when $\alpha\!\to\!0$, the generator leans toward user-tailored rewriting guided by $e_u$.

\subsection{Model Optimization}

PHG-DIF is optimized on the corpus $\mathcal{D}={(C_u,h_x,b_x,y)}$ according to the procedure detailed in Algorithm~\ref{alg:phg_dif}. 
We first pre-train the encoder $U_{\xi}$ on click-through labels, warm-up the decoder $G_{\theta}$ with \textit{maximum likelihood estimation}~(MLE), and fit the breaking-news classifier $B_{\psi}$ by \textit{binary cross-entropy}. 
The absence of personalized headline references makes it challenging to optimize personalized generation with purely supervised learning~\cite{song2023general}. Following PNG~\citep{ao2023put}, we therefore perform a policy-gradient fine-tuning step that maximizes the expected reward of sampled headlines.  
Specifically, the predicted breaking-news probability $\alpha$ is injected into the decoder gate to trade off personalization against factual fidelity, and an A2C search is adopted to estimate interim rewards.
The optimization objective is defined as:
\begin{align}
\mathcal{L}_{\text{RL}} &= -\mathbb{E}_{Y\sim G_{\theta}}\!\bigl[R(Y)\bigr], \label{eq:RL_loss}
\end{align}
where $R(Y)$ aggregates several headline-quality indicators.

\begin{algorithm}[t]
\caption{Model Optimization of PHG-DIF}
\label{alg:phg_dif}
\small
\KwIn{User Encoder $U_{\xi}$, Headline Generator $G_{\theta}$, Breaking-News Predictor $B_{\psi}$, dataset $\mathcal{D}$}
\KwOut{Optimized parameters $\xi,\,\theta,\,\psi$}
\BlankLine
\textbf{Randomly initialize} $\theta,\psi,\xi$\;
\BlankLine
\textbf{\# Phase 1: pre-train user encoder $U_{\xi}$}\;
\While{not converged}{
        Sample $(C_u)$ from $\mathcal{D}$\;
        Update $\xi$ via CTR prediction on $C_u$\;
}
\BlankLine
\textbf{\# Phase 2: MLE warm-up of headline generator $G_{\theta}$}\;
\While{not converged}{
        Sample $(C_u,b_x,h_x)$ from $\mathcal{D}$\;
        \textit{Freeze} $\xi$; compute $e_u=U_{\xi}(C_u)$, $v=\textsc{Enc}(b_x)$\;
        Update $\theta$ by maximising $\log P_{G_\theta}(h_x\mid e_u,v)$\;
}
\BlankLine
\textbf{\# Phase 3: train breaking-news predictor $B_{\psi}$}\;
\For{each $(b_x,y)$ in $\mathcal{D}$}{
        $v=\textsc{Enc}(b_x)$; update $\psi$ on $(v,y)$ with BCE loss\;
}
\BlankLine
\textbf{\# Phase 4: Policy-gradient fine-tuning (A2C)}\;
\While{not converged}{
        Sample $(C_u,b_x)$ from $\mathcal{D}$; $e_u\!\leftarrow\!U_{\xi}(C_u)$; $v\!\leftarrow\!\textsc{Enc}(b_x)$\;
        $\alpha\!\leftarrow\!B_{\psi}(v)$; $s_0\!\leftarrow\![\alpha e_u;\,v]$\;
        Generate headline $Y\!\sim\!G_{\theta}$; compute reward $R(Y)$\;
        Estimate advantage $\hat{A}_t$ and update $\theta,\xi$ via Eq.~\eqref{eq:RL_loss}\;
}
\Return{$\xi,\theta,\psi$}
\end{algorithm}

\section{Experimental Setup}
\subsection{DT-PENS Dataset}

We introduce DT-PENS, a personalized news headline generation benchmark annotated with user dwell times, addressing the lack of user historical dwell time data for offline evaluation in PENS~\cite{ao2021pens}. DT-PENS is a specialized dataset for personalized headline generation, featuring user dwell time annotations. DT-PENS is further developed in two phases from the anonymous user impressions in the training and validation sets of the original PENS.

\subsubsection{The First Phase}
We randomly sampled $1,000$ users from the PENS validation data. For each user, we extracted their detailed click history, corresponding article dwell and exposure times (where available), and news items that were exposed but not clicked. Subsequently, we leveraged Large Language Models (LLMs) to infer users' latent interests and generate preliminary personalized headlines based on these inferred interests and the candidate news. 
To ensure fairness, LLMs are not explicitly informed of the correlation between user click history and dwell time. Instead, we adopted a few-shot prompting strategy, which involved providing the models with partial anonymized historical interaction data and personalized headline samples derived from users in the original PENS test set, to guide the LLMs to learn and emulate the stylistic characteristics.
We generated over $40$K raw personalized headlines using multiple advanced LLMs, covering nearly $10$K candidate news. By incorporating a rejection sampling mechanism, we then preliminarily filtered these to obtain a substantial corpus of candidate personalized headlines exhibiting high initial quality.

\subsubsection{The Second Phase}
We designed a rigorous and meticulous multi-level filtering pipeline to ensure the quality and suitability of the final reference headlines. Firstly, we removed overly long or short headlines, ensuring their length distribution is comparable to personalized headlines in the original PENS dataset~\cite{ao2021pens}.
Furthermore, we eliminated headlines found to be irrelevant to the news articles, containing factual inaccuracies, or exhibiting potential hallucinations. To implement this, we computed the semantic similarity between each generated headline and its corresponding news article body. Headlines falling below a predefined similarity threshold were flagged as potentially irrelevant or hallucinatory and subsequently discarded. Finally, all candidate headlines that passed the aforementioned automated filtering stages were submitted to human annotators for a final review. 
For each test instance, human annotators identified the headline that best reflected the user's historical preferences, designating it as the ground truth. 
The final DT-PENS dataset consists of $9,823$ test instances from $1,000$ unique readers. 
Detailed procedures for the dataset's construction are provided in Appendix~\ref{appendix:imp}

\subsection{Baselines}
To comprehensively evaluate the performance of our proposed model, we select a diverse set of established baseline methods, encompassing both non-personalized and personalized methods.

The \textbf{non-personalized methods} include BART~\citep{lewis2019bart}, a bidirectional and autoregressive transformer model, and T5-small~\citep{raffel2020exploring}, a smaller variant of the T5 model designed for efficient text generation. These methods generate headlines without considering user preferences, serving as a general performance benchmark.

For \textbf{personalized methods}, we consider three personalized news headline generation methods that integrate user-specific preferences, namely PENS-EBNR, PENS-NRMS, and PENS-NAML, as mentioned by \citet{ao2021pens}.
Other personalized methods include PNG~\citep{ao2023put}, which tailors generated news headlines to individual users based on multi-perspective interests, and GTP~\citep{song2023general}, which improves personalized headlines through pre-training and achieved SOTA performance on the original PENS benchmark.

\subsection{Evaluation Metrics}
To ensure a fair comparison with previous studies on personalized news headline generation~\citet{ao2021pens, song2023general}, we evaluated the quality of generated personalized news headlines using several evaluation metrics.
For lexical similarity between the generated and reference headlines, we employ ROUGE-n~\citep{lin2004rouge}, which measures the overlap of n-grams and is widely used in text summarization evaluation\footnote{We use the rouge package provided by \url{https://github.com/pltrdy/rouge} for evaluation.}. 
To evaluate the semantic quality of the generated headlines, we utilize two model-based evaluation methods: BLEURT~\citep{sellam2020bleurt} and BARTScore~\citep{yuan2021bartscore}. BLEURT\footnote{\url{https://huggingface.co/spaces/evaluate-metric/bleurt}.} captures semantic similarity and provides robust quality judgments. BARTScore\footnote{\url{https://huggingface.co/ZoneTwelve/BARTScore}.} assesses fluency, grammar, and alignment with the input text by leveraging BART’s language understanding and generation probabilities.

\subsection{Implementation Details}
\label{subsec:ImplementationDetails}

To construct DT-PENS, we generated raw personalized headlines using multiple advanced LLMs, including o1-mini~\cite{jaech2024openai}, GPT-4o~\cite{hurst2024gpt}, GLM-4-plus~\cite{glm2024chatglm}, and the Qwen-2.5 series~\cite{qwen2.5}. All models were accessed via API endpoints with prompt engineering and a sampling temperature of $0.7$. The collected raw headlines were then scored automatically by Qwen-2.5-72B, which served as an LLM judge proxy. Raw headlines receiving extremely low scores were filtered out and resampled. The prompt templates and further details are provided in Appendix~\ref{appendix:imp}.
In the dual-filtering, we first selected the top $0.1\%$ of news articles in PENS~\cite{ao2021pens} based on click-through rate~(CTR), which formed the breaking news set for the news-level filtering. For time-level filtering, we set $k=30$ for the \textit{IPL}, applied a one-week sliding window for the \textit{IEA}, and determined the \textit{SIM} threshold based on the mean dwell time after excluding outliers~(dwell time $>3000$ s).
The encoder was implemented with 8-headed attention, while the decoder uses beam search with a beam width of $5$. The user model was pre-trained on a CTR prediction task, using a peak learning rate of $1e-5$. During PHG-DIF framework training, we applied a peak learning rate of $1e-7$ and executed an Advantage Actor-Critic (A2C) search with $16$ sampled sequences. We use the NVIDIA A800 80GB GPU for our experiments.

\begin{table*}[t!]
\centering
\setlength{\tabcolsep}{19pt}
\caption{Our main experimental results. ``\textit{-w/o}'' indicates component ablation (relative \% change in parentheses).}
\newcolumntype{N}{S[table-format=2.2, table-align-text-after=false]}
\begin{tabular}{l *{6}{N}} 
\toprule
\textbf{Methods} & \textbf{ROUGE-1} & \textbf{ROUGE-2} & \textbf{ROUGE-L} & \textbf{BLEURT} & \textbf{BARTScore} \\
\midrule
\multicolumn{7}{l}{\textbf{\textit{\ \textbullet }} \textbf{\textit{Non-personalized methods}}} \\
\addlinespace[1pt]
BART~\citep{lewis2019bart} & 18.05 & 6.70 & 17.19 & 26.36 & 58.03 \\
T5-small~\citep{raffel2020exploring} & 18.90 & 6.84 & 17.46 & 29.60 & 59.71 \\
\midrule
\multicolumn{7}{l}{\textbf{\textit{\ \textbullet }} \textbf{\textit{Personalized methods}}} \\
\addlinespace[1pt]
PENS-EBNR~\citep{ao2021pens} & 23.15 & 7.19 & 21.33 & 44.79 & 62.28 \\
PENS-NRMS~\citep{ao2021pens} & 22.89 & 7.01 & 21.15 & 43.40 & 60.27 \\
PENS-NAML~\citep{ao2021pens} & 23.11 & 7.17 & 21.21 & 44.18 & 62.20 \\
PNG~\citep{ao2023put} & 23.24 & 7.28 & 21.47 & 45.63 & 62.29 \\
GTP~\citep{song2023general} & 24.01 & 7.58 & 22.22 & 48.09 & 63.80 \\
\addlinespace[1pt]
\hdashline
\addlinespace[2pt]
$\textbf{PHG-DIF}_{\text{ours}}$ & {\bfseries 24.33} * & {\bfseries 7.99} * & {\bfseries 22.47} * & {\bfseries 48.50} * & {\bfseries 65.64} * \\
\textit{-w/o IPL} & 24.08 {\scriptsize{(-1.03\%)}} & 7.62 {\scriptsize{(-4.63\%)}} & 22.23 {\scriptsize{(-1.07\%)}} & 48.19 {\scriptsize{(-0.64\%)}} & 63.86 {\scriptsize{(-2.71\%)}} \\
\textit{-w/o IEA} & 24.16 {\scriptsize{(-0.70\%)}} & 7.64 {\scriptsize{(-4.38\%})} & 22.29 {\scriptsize{(-0.80\%)}} & 48.21 {\scriptsize{(-0.60\%)}} & 63.95 {\scriptsize{(-2.58\%)}} \\
\textit{-w/o SIM} & 23.65 {\scriptsize{(-2.80\%)}} & 7.32 {\scriptsize{(-8.39\%)}} & 21.76 {\scriptsize{(-3.16\%)}} & 46.43 {\scriptsize{(-4.27\%)}} & 62.78 {\scriptsize{(-4.36\%)}} \\
\bottomrule
\end{tabular}
\\[2pt]
\noindent \small
\textit{The symbol * denotes the significance level with $p \leq 0.05$. \textbf{Bold} font indicates the best-performing method.}
\label{main_res}
\end{table*}

\section{Results and Analysis}
In this section, we analyze our experiments to address the following research questions:

\begin{itemize}[leftmargin=1em]
  \item \textbf{RQ1:} How does PHG-DIF perform compared to competitive non-personalized and personalized headline generation baselines?
  \item \textbf{RQ2:} What factors affect the performance of PHG-DIF?
  \item \textbf{RQ3:} How do users perceive the personalized news headlines generated by PHG-DIF?
  \item \textbf{RQ4:} Why does PHG-DIF achieve these improvements?
\end{itemize}

\subsection{Overall Performance~(RQ1)}
To answer RQ1, we perform a comprehensive experimental evaluation of PHG-DIF. Table~\ref{main_res} reports the main results. Our proposed PHG-DIF framework outperforms all baseline methods across evaluation metrics, demonstrating that denoising users’ historical clickstreams markedly improve personalized headline generation. 

Compared to non-personalized methods, all personalized methods, including PHG-DIF, exhibit substantial gains as we expected. This finding underscores that personalization enhances headline quality by aligning with user preferences, highlighting the value of user-oriented strategies in news headline generation.

In fine-grained comparisons among personalized methods, we observe that the GTP achieves significantly higher ROUGE-L and BARTScore than earlier pointer-network-based approaches due to its strong pre-training backbone. This suggests that generic headline-generation pre-training yields beneficial effects for personalization, consistent with the observations of \citet{yang2023fact}. Nevertheless, our pointer-generator-based PHG-DIF still surpasses the GTP. We attribute this advantage to the dual-filtering mechanism in our user modeling, which is designed to effectively remove noise from users’ click histories. By applying this refined filtering, PHG-DIF distills a purer and more representative user interest profile from noisy interactions. 
These results indicate that in highly personalized contexts with complex user data, specialized noise filtering and dynamic interest modeling have the potential to surpass the generalization of pre-trained models.

\subsection{Ablation Study~(RQ2)}
\subsubsection{Impact of the Three Time-aware Encoders}
PHG-DIF embeds three time-aware encoders, detailed in Section \ref{subsec:time-encoder}. One important question that arises is how each of these time‐aware encoder modules contributes to the overall performance of PHG-DIF. To address this question, we conduct ablation studies on three variants, each omitting one encoder.
The results are shown in Table~\ref{main_res}, where ``\textit{-w/o}'' denotes the removal of the corresponding component. The results yield three observations:
1) All three encoders are essential. Removing IPL, IEA, or SIM individually causes noticeable drops on every evaluation metric, demonstrating that modeling instantaneous, evolving, and stable interests is critical.
2) Stable interest mining (SIM) has the greatest impact. Its removal causes the biggest degradation, highlighting the importance of long-term preference modeling.
3) Instantaneous preference learning (IPL) and interest evolution analysis (IEA) are complementary. While SIM is most influential, IPL and IEA are indispensable. Removing either produces moderate yet non-negligible losses, confirming the need to capture real-time and evolving interests.
Overall, the ablation results indicate that omitting any time-aware encoder degrades performance, thereby validating the effectiveness of the full design.

\subsubsection{Ablation on Breaking News Handling}

\begin{table}[tb]
\centering
\renewcommand{\arraystretch}{1.25}
\caption{Ablation study results for breaking news handling.}
\label{tab:ablation_breaking_news}
\resizebox{\linewidth}{!}{%
\begin{tabular}{lccccc}
  \toprule
  \textbf{Model} & \textbf{ROUGE-1} & \textbf{ROUGE-2} & \textbf{ROUGE-L} & \textbf{BLEURT} & \textbf{BARTScore} \\
  \midrule
PHG-DIF (full) & \textbf{24.33} & \textbf{7.99} & \textbf{22.47} & \textbf{48.50} & \textbf{65.64} \\
\textit{-w/o BF} & 24.12 & 7.63 & 22.25 & 48.20 & 63.91 \\
\textit{-w/o BP} & 24.29 & 7.82 & 22.40 & 48.41 & 65.16 \\
  \bottomrule
\end{tabular}
}
\end{table}

Breaking news is inherently compelling enough that readers will engage with it even without personalized headlines. Our approach therefore prioritizes factual accuracy for such news. We conducted an ablation study with two variants to examine the impact of removing either of the two breaking-news components in PHG-DIF. 
Table~\ref{tab:ablation_breaking_news} presents the results for two variants, where ``\textit{-w/o BF}” denotes removing the training-time \emph{Breaking News Filtering} (BF), ``\textit{-w/o BP}” denotes removing the inference-time \emph{Breaking Predictor} (BP).
When BF is ablated, performance drops markedly on every metric, confirming that news-level filtering is indispensable for eliminating collaborative popularity noise and for preserving genuine user signals in interest modeling. 
Conversely, omitting the BP component, which compels the generator to personalize all headlines, results in some factual breaking news headlines being replaced by less precise rewrites, consequently lowering semantic and n-gram scores. This finding indicates that BP accurately identifies breaking news at inference, allowing the model to adapt its generation strategy, thereby preserving factual accuracy for these items instead of invariably prioritizing personalization. The importance of safeguarding headline factuality for user experience, as evidenced by our results, aligns with the findings of \citet{yang2023fact}. 

\begin{figure}[b] 
    \centering 
    \begin{subfigure}[b]{0.48\linewidth} 
        \centering
        \includegraphics[width=\linewidth]{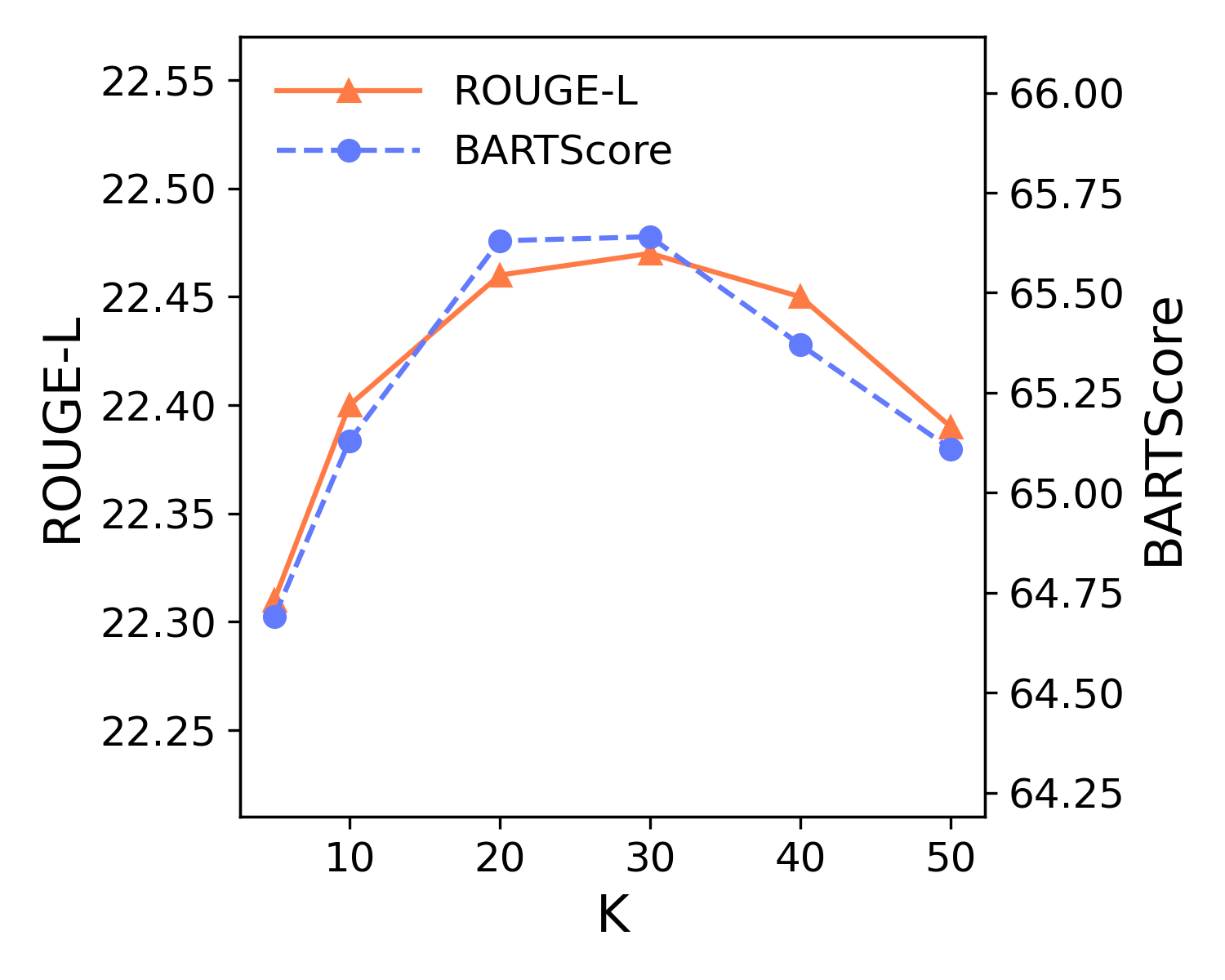}
        \caption{} 
        \vspace{-0.75em}
        \label{fig:k} 
    \end{subfigure}
    \hfill 
    \begin{subfigure}[b]{0.48\linewidth} 
        \centering
        \includegraphics[width=\linewidth]{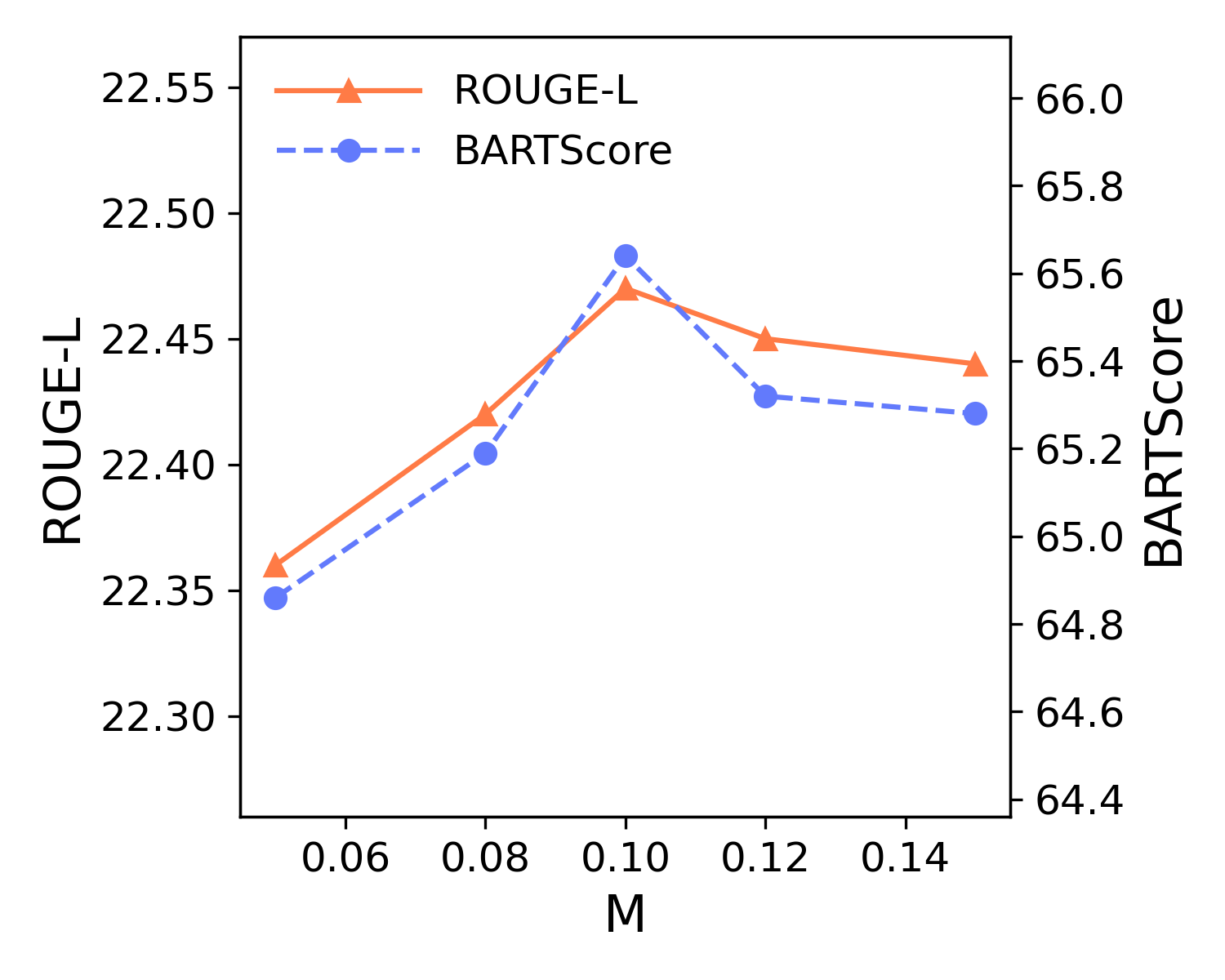} 
        \caption{} 
        \vspace{-0.75em}
        \label{fig:m} 
    \end{subfigure}
    \caption{Impact of IPL history length $K$ and breaking news threshold $M$ on model performance.} 
    \label{fig:hyper} 
\end{figure}

\subsubsection{Influence of History Length $K$ in IPL}
\label{subsec:ablation-k}
In this section, we investigate the impact of the historical click window size $K$ within the IPL module on overall model performance. As illustrated in Figure~\ref{fig:k}, our experimental results show that varying $K$ across the set $\{5, 10, 20, 30, 40, 50\}$ leads to a rapid improvement in model performance with increasing $K$, which peaks at $K=30$. 
We posit that an excessively small $K$ may prevent the model from capturing a sufficiently broad recent user history, resulting in inadequate modeling of instant preferences. Conversely, further increasing $K$ beyond this optimum might lead to functional overlap with modules designed for mid-to-long-term interests (e.g., IEA, SIM). This overlap could diminish IPL's distinct role in capturing short-term, immediate interests and, furthermore, potentially degrade performance due to information redundancy or introduced noise.

\subsubsection{Sensitivity to Threshold $M$}
\label{subsec:ablation-m}
Following GTP~\cite{song2023general}, we define breaking news as items ranking in \emph{the top-$M\%$} by click-through rate~(CTR). This definition aims to distinguish user clicks primarily driven by trending events or platform recommendations from those reflecting pure personal interest. 
A similar idea is also reflected in FPG~\cite{yang2023fact}, which restricts training to news items clicked by a limited number of users.
To ascertain the optimal value for $M$, we performed a sensitivity analysis on breaking news click-through rate threshold. We evaluated $M$ across a range from $0.05$ to $0.15$, affecting the label of $5,606,741$ news articles in the dataset. The results revealed optimal model performance at $M=0.10$~(corresponding to $0.10\%$ CTR threshold). We infer that an excessively low $M$ (e.g., $<0.10$) might be overly stringent, potentially leading to the omission of some breaking news that has garnered significant public attention. 
Conversely, an excessively high $M$ (e.g., $>0.10$) could cause a substantial volume of regular news to be misclassified as breaking news, which would then unnecessarily bypass the personalized rewriting process. 
This misclassification not only dilutes personalized user interest signals but also compromises the overall effectiveness of personalization. Hence, a moderate $M$ best balances popularity bias suppression, factual headlines for breaking news, and robust personalization elsewhere.

\subsection{User Study~(RQ3)}
\label{appendix:user}

\begin{table}[tb]
\centering
\small
\caption{Results of the user study with rankings.}
\resizebox{\linewidth}{!}{ 
\begin{tabular}{lccc}
\toprule
\textbf{Methods} & \textbf{Fluency} \ $\downarrow$ & \textbf{Consistency} \ $\downarrow$ & \textbf{Attractiveness} \ $\downarrow$ \\ 
\midrule
T5-small      & 2.95    & 3.02        & 3.25           \\
PNG           & 2.11    & 3.68        & 3.09           \\
GTP           & 2.84    & 1.68        & 1.95           \\
\hdashline
\addlinespace[2pt]
$\textbf{PHG-DIF}_{\text{ours}}$ & \textbf{2.10}    & \textbf{1.62}        & \textbf{1.77} \\ 
\bottomrule
\end{tabular}
}
\label{tab3}
\vspace{-1em}
\end{table}
To gain deeper insights into RQ3  and further evaluate the practical effectiveness of personalized news headlines generated by PHG-DIF, we conducted a user study. We recruited $5$ native English-speaking graduate students and compensated them according to our approved participant guidelines.
Participants were asked to select $100$ news articles from a preselected list, thereby constructing their user preference profiles. This process aimed to simulate their personalized historical clickstreams. Subsequently, four different personalized models generated headlines for $20$ unseen news articles based on each participant's historical click data. Participants assessed the generated headlines across three dimensions: \textbf{fluency}, \textbf{consistency}, and \textbf{attractiveness}, and ranked the headlines produced by each model~(with \textit{1 being the best} and \textit{4 the worst}). 
Notably, participants were unaware of the source of each headline and were allowed to assign the same ranking score to different headlines for the same news article. Finally, we calculated the average ranking of headlines generated by each model to obtain a ranking score, as shown in Table~\ref{tab3}. As observed, PHG-DIF achieved the highest scores in fluency, consistency, and attractiveness, suggesting that its personalized headlines are more aligned with users' true interests and have a greater potential to engage readers.

\subsection{Case Study~(RQ4)}
\label{appendix:case}

\begin{table*}[tbp]
\centering
\caption{A case on personalized headline generation affected by click noise. \textcolor{red}{Red text} highlights content related to click noise, and \textcolor{blue}{blue text} represents the user's true interests.}
\resizebox{0.85\linewidth}{!}{ 
\begin{tabular}{llc}
\toprule
\multicolumn{2}{l}{\textbf{Click History}} & \textbf{Dwell Time}\\[2pt]
\hline
\addlinespace[2pt]
\multicolumn{2}{l}{British Ambassador to the U.S., Kim Darroch, resigns after \textcolor{blue}{Trump} criticism} & 366s \\
\multicolumn{2}{l}{There’s a democratic civil war brewing over decriminalizing migration} & 115s \\
\multicolumn{2}{l}{A wooden sculpture of \textcolor{red}{Melania Trump} was unveiled on the banks of the Sava River} &  3s \\
\multicolumn{2}{l}{The true cost of high deductible health care plans} & 7s \\
\multicolumn{2}{l}{Report: \textcolor{blue}{Durant}, Irving planned to team up before 2018-19 season began} & 249s \\
\multicolumn{2}{l}{What were the \textcolor{red}{Warriors} thinking on \textcolor{red}{Stephen Curry}'s final shot?} & 5s \\
\multicolumn{2}{l}{The Knicks failed to sign \textcolor{blue}{Kevin Durant} and Kyrie Irving} & 486s \\
\midrule
\multicolumn{2}{l}{\textbf{\textbullet} \textit{\textbf{Case 1}}} & \\
\multicolumn{1}{c}{\textbf{Original Headline:}} & \multicolumn{2}{l}{Here’s when social security benefits could be cut} \\[2pt]
\hdashline
\addlinespace[2pt]
\textbf{PNG:} & Possible impact of \textcolor{red}{Melania Trump}’s social security reductions & \ding{55} \\
\textbf{GTP:} & The benefits of \textcolor{blue}{Trump}'s proposed social security reductions & \ding{51} \\
\textbf{Ours:} & Exploring implications of \textcolor{blue}{Trump}’s proposed social security cuts & \ding{51} \\
\midrule
\multicolumn{3}{l}{\textbf{\textbullet} \textit{\textbf{Case 2}}} \\
\textbf{Original Headline:} & \multicolumn{2}{l}{Calling BS on Stephen A. Smith's explosive claims on First Take} \\[2pt]
\hdashline
\addlinespace[2pt]
\textbf{PNG:} & \textcolor{red}{Stephen Curry} calls Stephen A.Smith's claims about the \textcolor{red}{Warriors} & \ding{55} \\
\textbf{GTP:} & \textcolor{blue}{Kevin Durant} addresses Stephen A. Smith's claims on his \textcolor{red}{Warriors} Era & \ding{55} \\
\textbf{Ours:} & \textcolor{blue}{Kevin Durant} denies Stephen A. Smith's claims about his departure & \ding{51} \\
\bottomrule
\end{tabular}
}
\label{tab2}
\end{table*}

To investigate why our PHG-DIF leads to improvements, we conducted a case study comparing the performance of the baseline PNG~\cite{ao2023put} and GTP~\cite{song2023general}. Table~\ref{tab2} outlines a user's click history, replete with dwell times that differentiate genuine interests from incidental clicks (i.e., click noise), alongside headlines generated by our proposed PHG-DIF and the baseline methods. The click history reveals a typical pattern where short-duration clicks represent click noise, contrasting with longer engagements that signify true user interests. Our analysis focuses on how effectively each model discerns these nuances. 
We found that the PNG erroneously interprets mistakenly clicked news as genuine user interests during the user modeling. This results in headlines that are both misaligned with user preferences and factually inaccurate. While the GTP produces headlines that are consistent with the news content, it still incorporates fake user interests. For instance, the generated headlines include information about the Warriors, despite the user's actual interest being in Kevin Durant and the Knicks. In contrast, our PHG-DIF method effectively filters out click noise, accurately capturing the user's true interests and generating headlines that better align with user preferences. The empirical case Study results underscore PHG-DIF's significant potential for enhancing user experience in real-world news recommendation systems, particularly in addressing the persistent challenge of click noise.

\section{Conclusion}
\label{sec:conclusion}
In this paper, we present PHG-DIF, a novel framework to tackle the challenges of click noise in user historical clickstreams. 
PHG-DIF employs a robust dual-filtering strategy that removes click noise at both news-level and time-level, isolating genuine user interests. Three specialized time-aware encoders then capture instantaneous, evolving, and stable preferences, yielding a precise user representation from noisy interaction data.
By denoising fake interests from implicit feedback, PHG-DIF effectively improves the precision of user profiles, leading to more relevant and accurate personalized headline generation. 
We further introduce DT-PENS, a new personalized headline generation benchmark with dwell time annotations for better evaluation. Extensive experiments on DT-PENS demonstrate that PHG-DIF significantly enhances headline quality and outperforms multiple competitive baseline methods.



\begin{acks}
The research work is supported by National Key R\&D Plan No. 2022YFC3303303, the National Natural Science Foundation of China under Grant (No. U2436209, 62476263), the Strategic Priority Research Program of the Chinese Academy of Sciences under Grant No. XDB0680201, Beijing Nova Program 20230484430, the Innovation Funding of ICT, CAS under Grant No. E461060.
\end{acks}

\appendix
\section{Limitations and Discussion}
Despite the significant improvements in headline quality and personalization achieved by our PHG-DIF framework for personalized news headline generation, we acknowledge several limitations that warrant further exploration in future research. 
Firstly, our PHG-DIF framework depends on historical clicks and dwell times. When interactions are sparse or noisy, the dual filtering module may not recover true interests, which can lower headline quality. The temporal fusion models gradual preference change yet may lag under abrupt shifts, such as breaking news. Our evaluation uses DT-PENS from a single platform with a limited cohort. Transfer to other platforms and user groups is uncertain. Cold start and missing click cases are underrepresented in current benchmarks. We recognize that the characteristics of this dataset may not be fully applicable to certain platforms or user groups. Specifically, for cases involving sparse or missing click behavior, such as cold-start scenarios, these issues extend beyond the scope of current PENS and DT-PENS benchmarks and are left for future work.

\section{DT-PENS Dataset Details}
\label{appendix:imp}

\begin{figure*}[t!]
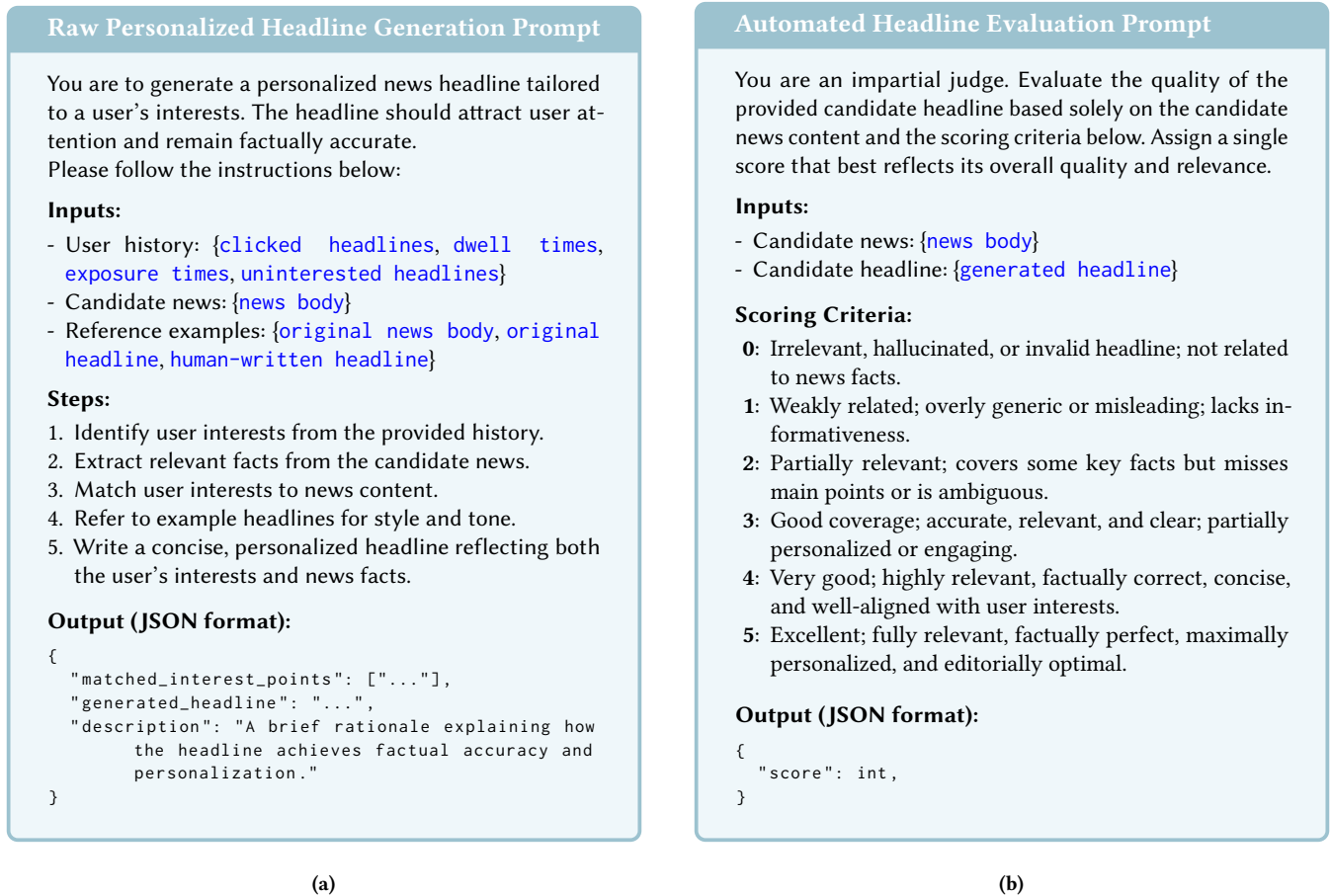
 
\centering 
\begin{subfigure}[t]{0.48\linewidth} 
\centering
\begin{tcolorbox}[
    colback=LightBlue!20,
    colframe=LightBlue!90!black,
    title=\large \textbf{Raw Personalized Headline Generation Prompt},
    width=\linewidth
]
\noindent
\textsf{You are to generate a personalized news headline tailored to a user's interests. The headline should attract user attention and remain factually accurate. }

\textsf{Please follow the instructions below:}

\vspace{0.5em}

\textsf{\textbf{Inputs:}}
\begin{itemize}[leftmargin=0.75em]
    \item[-] \textsf{User history: \{\textcolor{blue}{\texttt{clicked headlines}}, \textcolor{blue}{\texttt{dwell times}}, \textcolor{blue}{\texttt{exposure times}}, \textcolor{blue}{\texttt{uninterested headlines}}\}}
    \item[-] \textsf{Candidate news: \{\textcolor{blue}{\texttt{news body}}\}}
    \item[-] \textsf{Reference examples: \{\textcolor{blue}{\texttt{original news body}}, \textcolor{blue}{\texttt{original headline}}, \textcolor{blue}{\texttt{human-written headline}}\}}
\end{itemize}

\vspace{0.25em}

\textsf{\textbf{Steps:}}
\begin{enumerate}[label=\arabic*., leftmargin=*]
    \item \textsf{Identify user interests from the provided history.}
    \item \textsf{Extract relevant facts from the candidate news.}
    \item \textsf{Match user interests to news content.}
    \item \textsf{Refer to example headlines for style and tone.}
    \item \textsf{Write a concise, personalized headline reflecting both the user's interests and news facts.}
\end{enumerate}

\vspace{0.5em}

\textsf{\textbf{Output (JSON format):}}
\begin{lstlisting}[basicstyle=\ttfamily\footnotesize,breaklines=true]
{
  "matched_interest_points": ["..."],
  "generated_headline": "...",
  "description": "A brief rationale explaining how the headline achieves factual accuracy and personalization."
}
\end{lstlisting}

\end{tcolorbox}


\caption{} 
\label{fig:Generation_Prompt}
\end{subfigure}
\hfill 
\begin{subfigure}[t]{0.48\linewidth} 
\centering

\begin{tcolorbox}[
    colback=LightBlue!20,
    colframe=LightBlue!90!black,
    title=\large \textbf{Automated Headline Evaluation Prompt},  
    width=\linewidth
]
\noindent
\textsf{You are an impartial judge. Evaluate the quality of the provided candidate headline based solely on the candidate news content and the scoring criteria below. Assign a single score that best reflects its overall quality and relevance.}

\vspace{0.5em}

\textsf{\textbf{Inputs:}}
\begin{itemize}[leftmargin=0.75em]
    \item[-] \textsf{Candidate news: \{\textcolor{blue}{\texttt{news body}}\}}
    \item[-] \textsf{Candidate headline: \{\textcolor{blue}{\texttt{generated headline}}\}}
\end{itemize}

\vspace{0.5em}

\textsf{\textbf{Scoring Criteria:}}
\begin{itemize}[leftmargin=1.5em]
    \item[\textbf{0}:] Irrelevant, hallucinated, or invalid headline; not related to news facts.
    \item[\textbf{1}:] Weakly related; overly generic or misleading; lacks informativeness.
    \item[\textbf{2}:] Partially relevant; covers some key facts but misses main points or is ambiguous.
    \item[\textbf{3}:] Good coverage; accurate, relevant, and clear; partially personalized or engaging.
    \item[\textbf{4}:] Very good; highly relevant, factually correct, concise, and well-aligned with user interests.
    \item[\textbf{5}:] Excellent; fully relevant, factually perfect, maximally personalized, and editorially optimal.
\end{itemize}

\vspace{0.75em}

\textsf{\textbf{Output (JSON format):}}
\begin{lstlisting}[basicstyle=\ttfamily\footnotesize,breaklines=true]
{
  "score": int, 
}
\end{lstlisting}
\end{tcolorbox}


\caption{} 
\label{fig:Evaluation_Prompt}
\vspace{-0.5em}
\end{subfigure}
\caption{Prompt templates for instructing LLMs in the DT-PENS dataset construction.} 
\label{fig:Prompt_templates} 
\end{figure*}

This section further details the construction pipeline of the DT-PENS dataset.
To acquire raw personalized headlines from multiple advanced LLms, we employ a \emph{few-shot prompting} strategy~\cite{ouyang2022training}. We design a prompt template, as shown in Figure~\ref{fig:Generation_Prompt}, to guide LLMs in generating raw personalized headlines. The LLMs are provided with detailed historical click data for each user, including headlines of clicked news, dwell and exposure times, and unclicked headlines, along with few-shot examples from the original PENS test set. 
This setup enables LLMs to discern user interests and generate personalized headlines from the user's inferred perspective, aligning with their genuine interests.

Before forwarding over $40$K raw personalized headlines for final human vetting, we adopt the \emph{LLM-as-Judge} paradigm~\cite{li2024llms, dong2024can} to perform automatic scoring. 
We first devise a headline-quality rubric and instantiate it in a prompt template, as illustrated in Figure~\ref{fig:Evaluation_Prompt}. The judge LLM is then instructed to assign each headline an integer score from $0$ to $5$, where higher values indicate better headline quality. Headlines scoring below $2$ are discarded and re-sampled until the target corpus size is reached, substantially reducing the downstream manual workload. 

We post-process the outputs from the two-stage LLMs using a JSON parser, and any output that fails to parse is immediately discarded. All retained candidates are subsequently examined by human annotators. To guarantee objectivity, at least three annotators independently review each candidate. A candidate personalized headline is accepted only if at least two-thirds of the annotators concur that it meets the quality criteria.


\section*{GenAI Usage Disclosures}
We acknowledge the use of Generative AI (GenAI) tools in preparing this work. Specifically, we used multiple advanced LLMs to generate raw samples for the DT-PENS dataset (see Section~\ref{subsec:ImplementationDetails}), with all GenAI-generated content manually reviewed and validated. 
GenAI tools were also used for grammar checking and language refinement. We are fully responsible for all content and confirm this disclosure complies with ACM policy on GenAI use.

\bibliographystyle{ACM-Reference-Format}
\balance
\bibliography{sample-base-full}


\begin{thebibliography}{35}


\ifx \showCODEN    \undefined \def \showCODEN     #1{\unskip}     \fi
\ifx \showISBNx    \undefined \def \showISBNx     #1{\unskip}     \fi
\ifx \showISBNxiii \undefined \def \showISBNxiii  #1{\unskip}     \fi
\ifx \showISSN     \undefined \def \showISSN      #1{\unskip}     \fi
\ifx \showLCCN     \undefined \def \showLCCN      #1{\unskip}     \fi
\ifx \shownote     \undefined \def \shownote      #1{#1}          \fi
\ifx \showarticletitle \undefined \def \showarticletitle #1{#1}   \fi
\ifx \showURL      \undefined \def \showURL       {\relax}        \fi
\providecommand\bibfield[2]{#2}
\providecommand\bibinfo[2]{#2}
\providecommand\natexlab[1]{#1}
\providecommand\showeprint[2][]{arXiv:#2}

\bibitem[Ao et~al\mbox{.}(2023)]%
        {ao2023put}
\bibfield{author}{\bibinfo{person}{Xiang Ao}, \bibinfo{person}{Ling Luo}, \bibinfo{person}{Xiting Wang}, \bibinfo{person}{Zhao Yang}, \bibinfo{person}{Jiun-Hung Chen}, \bibinfo{person}{Ying Qiao}, \bibinfo{person}{Qing He}, {and} \bibinfo{person}{Xing Xie}.} \bibinfo{year}{2023}\natexlab{}.
\newblock \showarticletitle{Put Your Voice on Stage: Personalized Headline Generation for News Articles}.
\newblock \bibinfo{journal}{\emph{ACM Transactions on Knowledge Discovery from Data}} \bibinfo{volume}{18}, \bibinfo{number}{3} (\bibinfo{year}{2023}), \bibinfo{pages}{1--20}.
\newblock


\bibitem[Ao et~al\mbox{.}(2021)]%
        {ao2021pens}
\bibfield{author}{\bibinfo{person}{Xiang Ao}, \bibinfo{person}{Xiting Wang}, \bibinfo{person}{Ling Luo}, \bibinfo{person}{Ying Qiao}, \bibinfo{person}{Qing He}, {and} \bibinfo{person}{Xing Xie}.} \bibinfo{year}{2021}\natexlab{}.
\newblock \showarticletitle{PENS: A dataset and generic framework for personalized news headline generation}. In \bibinfo{booktitle}{\emph{Proceedings of the 59th Annual Meeting of the Association for Computational Linguistics and the 11th International Joint Conference on Natural Language Processing (Volume 1: Long Papers)}}. \bibinfo{pages}{82--92}.
\newblock


\bibitem[Devlin et~al\mbox{.}(2019)]%
        {devlin-etal-2019-bert}
\bibfield{author}{\bibinfo{person}{Jacob Devlin}, \bibinfo{person}{Ming-Wei Chang}, \bibinfo{person}{Kenton Lee}, {and} \bibinfo{person}{Kristina Toutanova}.} \bibinfo{year}{2019}\natexlab{}.
\newblock \showarticletitle{{BERT}: Pre-training of Deep Bidirectional Transformers for Language Understanding}. In \bibinfo{booktitle}{\emph{Proceedings of the 2019 Conference of the North {A}merican Chapter of the Association for Computational Linguistics: Human Language Technologies, Volume 1 (Long and Short Papers)}}, \bibfield{editor}{\bibinfo{person}{Jill Burstein}, \bibinfo{person}{Christy Doran}, {and} \bibinfo{person}{Thamar Solorio}} (Eds.). \bibinfo{publisher}{Association for Computational Linguistics}, \bibinfo{address}{Minneapolis, Minnesota}, \bibinfo{pages}{4171--4186}.
\newblock
\href{https://doi.org/10.18653/v1/N19-1423}{doi:\nolinkurl{10.18653/v1/N19-1423}}


\bibitem[Dong et~al\mbox{.}(2024)]%
        {dong2024can}
\bibfield{author}{\bibinfo{person}{Yijiang~River Dong}, \bibinfo{person}{Tiancheng Hu}, {and} \bibinfo{person}{Nigel Collier}.} \bibinfo{year}{2024}\natexlab{}.
\newblock \showarticletitle{Can LLM be a Personalized Judge?}
\newblock \bibinfo{journal}{\emph{arXiv preprint arXiv:2406.11657}} (\bibinfo{year}{2024}).
\newblock


\bibitem[GLM et~al\mbox{.}(2024)]%
        {glm2024chatglm}
\bibfield{author}{\bibinfo{person}{Team GLM}, \bibinfo{person}{Aohan Zeng}, \bibinfo{person}{Bin Xu}, \bibinfo{person}{Bowen Wang}, \bibinfo{person}{Chenhui Zhang}, \bibinfo{person}{Da Yin}, \bibinfo{person}{Dan Zhang}, \bibinfo{person}{Diego Rojas}, \bibinfo{person}{Guanyu Feng}, \bibinfo{person}{Hanlin Zhao}, {et~al\mbox{.}}} \bibinfo{year}{2024}\natexlab{}.
\newblock \showarticletitle{Chatglm: A family of large language models from glm-130b to glm-4 all tools}.
\newblock \bibinfo{journal}{\emph{arXiv preprint arXiv:2406.12793}} (\bibinfo{year}{2024}).
\newblock


\bibitem[Hurst et~al\mbox{.}(2024)]%
        {hurst2024gpt}
\bibfield{author}{\bibinfo{person}{Aaron Hurst}, \bibinfo{person}{Adam Lerer}, \bibinfo{person}{Adam~P Goucher}, \bibinfo{person}{Adam Perelman}, \bibinfo{person}{Aditya Ramesh}, \bibinfo{person}{Aidan Clark}, \bibinfo{person}{AJ Ostrow}, \bibinfo{person}{Akila Welihinda}, \bibinfo{person}{Alan Hayes}, \bibinfo{person}{Alec Radford}, {et~al\mbox{.}}} \bibinfo{year}{2024}\natexlab{}.
\newblock \showarticletitle{Gpt-4o system card}.
\newblock \bibinfo{journal}{\emph{arXiv preprint arXiv:2410.21276}} (\bibinfo{year}{2024}).
\newblock


\bibitem[Jaech et~al\mbox{.}(2024)]%
        {jaech2024openai}
\bibfield{author}{\bibinfo{person}{Aaron Jaech}, \bibinfo{person}{Adam Kalai}, \bibinfo{person}{Adam Lerer}, \bibinfo{person}{Adam Richardson}, \bibinfo{person}{Ahmed El-Kishky}, \bibinfo{person}{Aiden Low}, \bibinfo{person}{Alec Helyar}, \bibinfo{person}{Aleksander Madry}, \bibinfo{person}{Alex Beutel}, \bibinfo{person}{Alex Carney}, {et~al\mbox{.}}} \bibinfo{year}{2024}\natexlab{}.
\newblock \showarticletitle{Openai o1 system card}.
\newblock \bibinfo{journal}{\emph{arXiv preprint arXiv:2412.16720}} (\bibinfo{year}{2024}).
\newblock


\bibitem[Jiang et~al\mbox{.}(2024)]%
        {jiang2024time}
\bibfield{author}{\bibinfo{person}{Hao Jiang}, \bibinfo{person}{Chuanzhen Li}, {and} \bibinfo{person}{Mingxiao An}.} \bibinfo{year}{2024}\natexlab{}.
\newblock \showarticletitle{Time Matters: Enhancing Pre-trained News Recommendation Models with Robust User Dwell Time Injection}.
\newblock \bibinfo{journal}{\emph{arXiv preprint arXiv:2405.12486}} (\bibinfo{year}{2024}).
\newblock


\bibitem[Kim and Chan(2003)]%
        {kim2003learning}
\bibfield{author}{\bibinfo{person}{Hyoung~R Kim} {and} \bibinfo{person}{Philip~K Chan}.} \bibinfo{year}{2003}\natexlab{}.
\newblock \showarticletitle{Learning implicit user interest hierarchy for context in personalization}. In \bibinfo{booktitle}{\emph{Proceedings of the 8th international conference on Intelligent user interfaces}}. \bibinfo{pages}{101--108}.
\newblock


\bibitem[Lewis et~al\mbox{.}(2020)]%
        {lewis2019bart}
\bibfield{author}{\bibinfo{person}{Mike Lewis}, \bibinfo{person}{Yinhan Liu}, \bibinfo{person}{Naman Goyal}, \bibinfo{person}{Marjan Ghazvininejad}, \bibinfo{person}{Abdelrahman Mohamed}, \bibinfo{person}{Omer Levy}, \bibinfo{person}{Veselin Stoyanov}, {and} \bibinfo{person}{Luke Zettlemoyer}.} \bibinfo{year}{2020}\natexlab{}.
\newblock \showarticletitle{BART: Denoising Sequence-to-Sequence Pre-training for Natural Language Generation, Translation, and Comprehension.}. In \bibinfo{booktitle}{\emph{Proceedings of the 58th Annual Meeting of the Association for Computational Linguistics}}.
\newblock
\href{https://doi.org/10.18653/v1/2020.acl-main.703}{doi:\nolinkurl{10.18653/v1/2020.acl-main.703}}


\bibitem[Li et~al\mbox{.}(2024)]%
        {li2024llms}
\bibfield{author}{\bibinfo{person}{Haitao Li}, \bibinfo{person}{Qian Dong}, \bibinfo{person}{Junjie Chen}, \bibinfo{person}{Huixue Su}, \bibinfo{person}{Yujia Zhou}, \bibinfo{person}{Qingyao Ai}, \bibinfo{person}{Ziyi Ye}, {and} \bibinfo{person}{Yiqun Liu}.} \bibinfo{year}{2024}\natexlab{}.
\newblock \showarticletitle{Llms-as-judges: a comprehensive survey on llm-based evaluation methods}.
\newblock \bibinfo{journal}{\emph{arXiv preprint arXiv:2412.05579}} (\bibinfo{year}{2024}).
\newblock


\bibitem[Li et~al\mbox{.}(2022)]%
        {li2022news}
\bibfield{author}{\bibinfo{person}{Zhengpeng Li}, \bibinfo{person}{Jiansheng Wu}, \bibinfo{person}{Jiawei Miao}, {and} \bibinfo{person}{Xinmiao Yu}.} \bibinfo{year}{2022}\natexlab{}.
\newblock \showarticletitle{News headline generation based on improved decoder from transformer}.
\newblock \bibinfo{journal}{\emph{Scientific Reports}} \bibinfo{volume}{12}, \bibinfo{number}{1} (\bibinfo{year}{2022}), \bibinfo{pages}{11648}.
\newblock


\bibitem[Lian et~al\mbox{.}(2025)]%
        {lian2025panoramic}
\bibfield{author}{\bibinfo{person}{Junhong Lian}, \bibinfo{person}{Xiang Ao}, \bibinfo{person}{Xinyu Liu}, \bibinfo{person}{Yang Liu}, {and} \bibinfo{person}{Qing He}.} \bibinfo{year}{2025}\natexlab{}.
\newblock \showarticletitle{Panoramic Interests: Stylistic-Content Aware Personalized Headline Generation}. In \bibinfo{booktitle}{\emph{Companion Proceedings of the ACM on Web Conference 2025}}. \bibinfo{pages}{1109--1112}.
\newblock


\bibitem[Lin(2004)]%
        {lin2004rouge}
\bibfield{author}{\bibinfo{person}{Chin-Yew Lin}.} \bibinfo{year}{2004}\natexlab{}.
\newblock \showarticletitle{Rouge: A package for automatic evaluation of summaries}. In \bibinfo{booktitle}{\emph{Text summarization branches out}}. \bibinfo{pages}{74--81}.
\newblock


\bibitem[Luo et~al\mbox{.}(2019)]%
        {luo2019reading}
\bibfield{author}{\bibinfo{person}{Ling Luo}, \bibinfo{person}{Xiang Ao}, \bibinfo{person}{Yan Song}, \bibinfo{person}{Feiyang Pan}, \bibinfo{person}{Min Yang}, {and} \bibinfo{person}{Qing He}.} \bibinfo{year}{2019}\natexlab{}.
\newblock \showarticletitle{Reading like HER: Human reading inspired extractive summarization}. In \bibinfo{booktitle}{\emph{Proceedings of the 2019 Conference on Empirical Methods in Natural Language Processing and the 9th International Joint Conference on Natural Language Processing (EMNLP-IJCNLP)}}. \bibinfo{pages}{3033--3043}.
\newblock


\bibitem[Nallapati et~al\mbox{.}(2016)]%
        {nallapati2016abstractive}
\bibfield{author}{\bibinfo{person}{Ramesh Nallapati}, \bibinfo{person}{Bowen Zhou}, \bibinfo{person}{Cicero Nogueira~Dos Santos}, \bibinfo{person}{Caglar Gulcehre}, {and} \bibinfo{person}{Bing Xiang}.} \bibinfo{year}{2016}\natexlab{}.
\newblock \showarticletitle{Abstractive Text Summarization using Sequence-to-sequence RNNs and Beyond}. In \bibinfo{booktitle}{\emph{Proceedings of the 20th SIGNLL Conference on Computational Natural Language Learning}}. \bibinfo{pages}{280--290}.
\newblock


\bibitem[Ouyang et~al\mbox{.}(2022)]%
        {ouyang2022training}
\bibfield{author}{\bibinfo{person}{Long Ouyang}, \bibinfo{person}{Jeffrey Wu}, \bibinfo{person}{Xu Jiang}, \bibinfo{person}{Diogo Almeida}, \bibinfo{person}{Carroll Wainwright}, \bibinfo{person}{Pamela Mishkin}, \bibinfo{person}{Chong Zhang}, \bibinfo{person}{Sandhini Agarwal}, \bibinfo{person}{Katarina Slama}, \bibinfo{person}{Alex Ray}, {et~al\mbox{.}}} \bibinfo{year}{2022}\natexlab{}.
\newblock \showarticletitle{Training language models to follow instructions with human feedback}.
\newblock \bibinfo{journal}{\emph{Advances in neural information processing systems}}  \bibinfo{volume}{35} (\bibinfo{year}{2022}), \bibinfo{pages}{27730--27744}.
\newblock


\bibitem[Qi et~al\mbox{.}(2021)]%
        {qi2021hierec}
\bibfield{author}{\bibinfo{person}{Tao Qi}, \bibinfo{person}{Fangzhao Wu}, \bibinfo{person}{Chuhan Wu}, \bibinfo{person}{Peiru Yang}, \bibinfo{person}{Yang Yu}, \bibinfo{person}{Xing Xie}, {and} \bibinfo{person}{Yongfeng Huang}.} \bibinfo{year}{2021}\natexlab{}.
\newblock \showarticletitle{HieRec: Hierarchical User Interest Modeling for Personalized News Recommendation}. In \bibinfo{booktitle}{\emph{Proceedings of the 59th Annual Meeting of the Association for Computational Linguistics and the 11th International Joint Conference on Natural Language Processing (Volume 1: Long Papers)}}. \bibinfo{pages}{5446--5456}.
\newblock


\bibitem[Qian et~al\mbox{.}(2023)]%
        {qian2023hutcrs}
\bibfield{author}{\bibinfo{person}{Mingjie Qian}, \bibinfo{person}{Yongsen Zheng}, \bibinfo{person}{Jinghui Qin}, {and} \bibinfo{person}{Liang Lin}.} \bibinfo{year}{2023}\natexlab{}.
\newblock \showarticletitle{HutCRS: Hierarchical user-interest tracking for conversational recommender system}. In \bibinfo{booktitle}{\emph{Proceedings of the 2023 Conference on Empirical Methods in Natural Language Processing}}. \bibinfo{pages}{10281--10290}.
\newblock


\bibitem[Raffel et~al\mbox{.}(2020)]%
        {raffel2020exploring}
\bibfield{author}{\bibinfo{person}{Colin Raffel}, \bibinfo{person}{Noam Shazeer}, \bibinfo{person}{Adam Roberts}, \bibinfo{person}{Katherine Lee}, \bibinfo{person}{Sharan Narang}, \bibinfo{person}{Michael Matena}, \bibinfo{person}{Yanqi Zhou}, \bibinfo{person}{Wei Li}, {and} \bibinfo{person}{Peter~J Liu}.} \bibinfo{year}{2020}\natexlab{}.
\newblock \showarticletitle{Exploring the limits of transfer learning with a unified text-to-text transformer}.
\newblock \bibinfo{journal}{\emph{Journal of machine learning research}} \bibinfo{volume}{21}, \bibinfo{number}{140} (\bibinfo{year}{2020}), \bibinfo{pages}{1--67}.
\newblock
\urldef\tempurl%
\url{http://jmlr.org/papers/v21/20-074.html}
\showURL{%
\tempurl}


\bibitem[Rayner et~al\mbox{.}(2016)]%
        {rayner2016so}
\bibfield{author}{\bibinfo{person}{Keith Rayner}, \bibinfo{person}{Elizabeth~R Schotter}, \bibinfo{person}{Michael~EJ Masson}, \bibinfo{person}{Mary~C Potter}, {and} \bibinfo{person}{Rebecca Treiman}.} \bibinfo{year}{2016}\natexlab{}.
\newblock \showarticletitle{So much to read, so little time: How do we read, and can speed reading help?}
\newblock \bibinfo{journal}{\emph{Psychological Science in the Public Interest}} \bibinfo{volume}{17}, \bibinfo{number}{1} (\bibinfo{year}{2016}), \bibinfo{pages}{4--34}.
\newblock


\bibitem[Rush et~al\mbox{.}(2015)]%
        {rush2015neural}
\bibfield{author}{\bibinfo{person}{Alexander~M Rush}, \bibinfo{person}{Sumit Chopra}, {and} \bibinfo{person}{Jason Weston}.} \bibinfo{year}{2015}\natexlab{}.
\newblock \showarticletitle{A Neural Attention Model for Abstractive Sentence Summarization}. In \bibinfo{booktitle}{\emph{Proceedings of the 2015 Conference on Empirical Methods in Natural Language Processing}}. \bibinfo{pages}{379--389}.
\newblock


\bibitem[Salemi et~al\mbox{.}(2024)]%
        {salemi2024lamp}
\bibfield{author}{\bibinfo{person}{Alireza Salemi}, \bibinfo{person}{Sheshera Mysore}, \bibinfo{person}{Michael Bendersky}, {and} \bibinfo{person}{Hamed Zamani}.} \bibinfo{year}{2024}\natexlab{}.
\newblock \showarticletitle{LaMP: When Large Language Models Meet Personalization}. In \bibinfo{booktitle}{\emph{Proceedings of the 62nd Annual Meeting of the Association for Computational Linguistics (Volume 1: Long Papers)}}. \bibinfo{pages}{7370--7392}.
\newblock


\bibitem[See et~al\mbox{.}(2017)]%
        {See_Liu_Manning_2017}
\bibfield{author}{\bibinfo{person}{Abigail See}, \bibinfo{person}{Peter~J. Liu}, {and} \bibinfo{person}{Christopher~D. Manning}.} \bibinfo{year}{2017}\natexlab{}.
\newblock \showarticletitle{Get To The Point: Summarization with Pointer-Generator Networks}. In \bibinfo{booktitle}{\emph{Proceedings of the 55th Annual Meeting of the Association for Computational Linguistics (Volume 1: Long Papers)}}.
\newblock
\href{https://doi.org/10.18653/v1/p17-1099}{doi:\nolinkurl{10.18653/v1/p17-1099}}


\bibitem[Sellam et~al\mbox{.}(2020)]%
        {sellam2020bleurt}
\bibfield{author}{\bibinfo{person}{Thibault Sellam}, \bibinfo{person}{Dipanjan Das}, {and} \bibinfo{person}{Ankur Parikh}.} \bibinfo{year}{2020}\natexlab{}.
\newblock \showarticletitle{BLEURT: Learning Robust Metrics for Text Generation}. In \bibinfo{booktitle}{\emph{Proceedings of the 58th Annual Meeting of the Association for Computational Linguistics}}. \bibinfo{pages}{7881--7892}.
\newblock


\bibitem[Song et~al\mbox{.}(2023)]%
        {song2023general}
\bibfield{author}{\bibinfo{person}{Yun-Zhu Song}, \bibinfo{person}{Yi-Syuan Chen}, \bibinfo{person}{Lu Wang}, {and} \bibinfo{person}{Hong-Han Shuai}.} \bibinfo{year}{2023}\natexlab{}.
\newblock \showarticletitle{General then Personal: Decoupling and Pre-training for Personalized Headline Generation}.
\newblock \bibinfo{journal}{\emph{Transactions of the Association for Computational Linguistics}}  \bibinfo{volume}{11} (\bibinfo{year}{2023}), \bibinfo{pages}{1588--1607}.
\newblock


\bibitem[Tan et~al\mbox{.}(2024)]%
        {tan2024enhancing}
\bibfield{author}{\bibinfo{person}{Xiaoyu Tan}, \bibinfo{person}{Leijun Cheng}, \bibinfo{person}{Xihe Qiu}, \bibinfo{person}{Shaojie Shi}, \bibinfo{person}{Yuan Cheng}, \bibinfo{person}{Wei Chu}, \bibinfo{person}{Yinghui Xu}, {and} \bibinfo{person}{Yuan Qi}.} \bibinfo{year}{2024}\natexlab{}.
\newblock \showarticletitle{Enhancing Personalized Headline Generation via Offline Goal-conditioned Reinforcement Learning with Large Language Models}. In \bibinfo{booktitle}{\emph{Proceedings of the 30th ACM SIGKDD Conference on Knowledge Discovery and Data Mining}}. \bibinfo{pages}{5762--5772}.
\newblock


\bibitem[Wang et~al\mbox{.}(2021)]%
        {wang2021denoising}
\bibfield{author}{\bibinfo{person}{Wenjie Wang}, \bibinfo{person}{Fuli Feng}, \bibinfo{person}{Xiangnan He}, \bibinfo{person}{Liqiang Nie}, {and} \bibinfo{person}{Tat-Seng Chua}.} \bibinfo{year}{2021}\natexlab{}.
\newblock \showarticletitle{Denoising implicit feedback for recommendation}. In \bibinfo{booktitle}{\emph{Proceedings of the 14th ACM international conference on web search and data mining}}. \bibinfo{pages}{373--381}.
\newblock


\bibitem[Xie et~al\mbox{.}(2021)]%
        {xie2021deep}
\bibfield{author}{\bibinfo{person}{Ruobing Xie}, \bibinfo{person}{Cheng Ling}, \bibinfo{person}{Yalong Wang}, \bibinfo{person}{Rui Wang}, \bibinfo{person}{Feng Xia}, {and} \bibinfo{person}{Leyu Lin}.} \bibinfo{year}{2021}\natexlab{}.
\newblock \showarticletitle{Deep feedback network for recommendation}. In \bibinfo{booktitle}{\emph{Proceedings of the twenty-ninth international conference on international joint conferences on artificial intelligence}}. \bibinfo{pages}{2519--2525}.
\newblock


\bibitem[Xie et~al\mbox{.}(2023)]%
        {xie2023reweighting}
\bibfield{author}{\bibinfo{person}{Ruobing Xie}, \bibinfo{person}{Lin Ma}, \bibinfo{person}{Shaoliang Zhang}, \bibinfo{person}{Feng Xia}, {and} \bibinfo{person}{Leyu Lin}.} \bibinfo{year}{2023}\natexlab{}.
\newblock \showarticletitle{Reweighting Clicks with Dwell Time in Recommendation}. In \bibinfo{booktitle}{\emph{Companion Proceedings of the ACM Web Conference 2023}}. \bibinfo{pages}{341--345}.
\newblock


\bibitem[Yang et~al\mbox{.}(2024)]%
        {qwen2.5}
\bibfield{author}{\bibinfo{person}{An Yang}, \bibinfo{person}{Baosong Yang}, \bibinfo{person}{Beichen Zhang}, \bibinfo{person}{Binyuan Hui}, \bibinfo{person}{Bo Zheng}, \bibinfo{person}{Bowen Yu}, \bibinfo{person}{Chengyuan Li}, \bibinfo{person}{Dayiheng Liu}, \bibinfo{person}{Fei Huang}, \bibinfo{person}{Haoran Wei}, \bibinfo{person}{Huan Lin}, \bibinfo{person}{Jian Yang}, \bibinfo{person}{Jianhong Tu}, \bibinfo{person}{Jianwei Zhang}, \bibinfo{person}{Jianxin Yang}, \bibinfo{person}{Jiaxi Yang}, \bibinfo{person}{Jingren Zhou}, \bibinfo{person}{Junyang Lin}, \bibinfo{person}{Kai Dang}, \bibinfo{person}{Keming Lu}, \bibinfo{person}{Keqin Bao}, \bibinfo{person}{Kexin Yang}, \bibinfo{person}{Le Yu}, \bibinfo{person}{Mei Li}, \bibinfo{person}{Mingfeng Xue}, \bibinfo{person}{Pei Zhang}, \bibinfo{person}{Qin Zhu}, \bibinfo{person}{Rui Men}, \bibinfo{person}{Runji Lin}, \bibinfo{person}{Tianhao Li}, \bibinfo{person}{Tingyu Xia}, \bibinfo{person}{Xingzhang Ren}, \bibinfo{person}{Xuancheng Ren}, \bibinfo{person}{Yang Fan}, \bibinfo{person}{Yang Su}, \bibinfo{person}{Yichang Zhang}, \bibinfo{person}{Yu Wan}, \bibinfo{person}{Yuqiong Liu}, \bibinfo{person}{Zeyu Cui}, \bibinfo{person}{Zhenru Zhang}, {and} \bibinfo{person}{Zihan Qiu}.} \bibinfo{year}{2024}\natexlab{}.
\newblock \showarticletitle{Qwen2.5 Technical Report}.
\newblock \bibinfo{journal}{\emph{arXiv preprint arXiv:2412.15115}} (\bibinfo{year}{2024}).
\newblock


\bibitem[Yang et~al\mbox{.}(2023)]%
        {yang2023fact}
\bibfield{author}{\bibinfo{person}{Zhao Yang}, \bibinfo{person}{Junhong Lian}, {and} \bibinfo{person}{Xiang Ao}.} \bibinfo{year}{2023}\natexlab{}.
\newblock \showarticletitle{Fact-Preserved Personalized News Headline Generation}. In \bibinfo{booktitle}{\emph{2023 IEEE International Conference on Data Mining (ICDM)}}. IEEE, \bibinfo{pages}{1493--1498}.
\newblock


\bibitem[Yuan et~al\mbox{.}(2021)]%
        {yuan2021bartscore}
\bibfield{author}{\bibinfo{person}{Weizhe Yuan}, \bibinfo{person}{Graham Neubig}, {and} \bibinfo{person}{Pengfei Liu}.} \bibinfo{year}{2021}\natexlab{}.
\newblock \showarticletitle{Bartscore: Evaluating generated text as text generation}.
\newblock \bibinfo{journal}{\emph{Advances in Neural Information Processing Systems}}  \bibinfo{volume}{34} (\bibinfo{year}{2021}), \bibinfo{pages}{27263--27277}.
\newblock


\bibitem[Zhang et~al\mbox{.}(2022)]%
        {zhang2022personalized}
\bibfield{author}{\bibinfo{person}{Kui Zhang}, \bibinfo{person}{Guangquan Lu}, \bibinfo{person}{Guixian Zhang}, \bibinfo{person}{Zhi Lei}, {and} \bibinfo{person}{Lijuan Wu}.} \bibinfo{year}{2022}\natexlab{}.
\newblock \showarticletitle{Personalized headline generation with enhanced user interest perception}. In \bibinfo{booktitle}{\emph{International Conference on Artificial Neural Networks}}. Springer, \bibinfo{pages}{797--809}.
\newblock


\bibitem[Zhou et~al\mbox{.}(2019)]%
        {zhou2019deep}
\bibfield{author}{\bibinfo{person}{Guorui Zhou}, \bibinfo{person}{Na Mou}, \bibinfo{person}{Ying Fan}, \bibinfo{person}{Qi Pi}, \bibinfo{person}{Weijie Bian}, \bibinfo{person}{Chang Zhou}, \bibinfo{person}{Xiaoqiang Zhu}, {and} \bibinfo{person}{Kun Gai}.} \bibinfo{year}{2019}\natexlab{}.
\newblock \showarticletitle{Deep interest evolution network for click-through rate prediction}. In \bibinfo{booktitle}{\emph{Proceedings of the AAAI conference on artificial intelligence}}, Vol.~\bibinfo{volume}{33}. \bibinfo{pages}{5941--5948}.
\newblock


\end{thebibliography}

\end{document}